# Online Learning of Assignments that Maximize Submodular Functions

**Daniel Golovin**
Carnegie Mellon University
Pittsburgh, PA 15213
dgolovin@cs.cmu.edu

**Andreas Krause**
California Institute of Technology
Pasadena, CA 91125
krausea@caltech.edu

**Matthew Streeter**
Google, Inc.
Pittsburgh, PA 15213
mstreeter@google.com

## Abstract

Which ads should we display in sponsored search in order to maximize our revenue? How should we dynamically rank information sources to maximize value of information? These applications exhibit strong diminishing returns: Selection of redundant ads and information sources decreases their marginal utility. We show that these and other problems can be formalized as repeatedly selecting an assignment of items to positions to maximize a sequence of *monotone submodular* functions that arrive one by one. We present an efficient algorithm for this general problem and analyze it in the *no-regret* model. Our algorithm possesses strong theoretical guarantees, such as a performance ratio that converges to the optimal constant of $1 - 1/e$. We empirically evaluate our algorithm on two real-world online optimization problems on the web: ad allocation with submodular utilities, and dynamically ranking blogs to detect information cascades.

## 1 Introduction

Consider the problem of repeatedly choosing advertisements to display in sponsored search to maximize our revenue. In this problem, there is a small set of positions on the page, and each time a query arrives we would like to assign, to each position, one out of a large number of possible ads. In this and related problems that we call *online assignment learning* problems, there is a set of positions, a set of items, and a sequence of rounds, and on each round we must assign an item to each position. After each round, we obtain some reward depending on the selected assignment, and we observe the value of the reward. When there is only one position, this problem becomes the well-studied multiarmed bandit problem [2]. When the positions have a linear ordering the assignment can be construed as a ranked list of elements, and the problem becomes one of selecting lists online. Online assignment learning thus models a central challenge in web search, sponsored search, news aggregators, and recommendation systems, among other applications.

A common assumption made in previous work on these problems is that the quality of an assignment is the sum of a function on the (item, position) pairs in the assignment. For example, online advertising models with *click-through-rates* [6] make an assumption of this form. More recently, there have been attempts to incorporate the value of diversity in the reward function [16]. Intuitively, even though the best $K$ results for the query "turkey" might happen to be about the country, the best list of $K$ results is likely to contain some recipes for the bird as well. This will be the case if there are diminishing returns on the number of relevant links presented to a user; for example, if it is better to present each user with at least one relevant result than to present half of the users with no relevant results and half with two relevant results. We incorporate these considerations in a flexible way by providing an algorithm that performs well whenever the reward for an assignment is a *monotone submodular function* of its set of (item, position) pairs.

Our key contributions are: (*i*) an efficient algorithm, TABULARGREEDY, that provides a constant factor $(1 - 1/e)$ approximation to the problem of optimizing assignments under submodular utility functions, (*ii*) an algorithm for online learning of assignments, TGBANDIT, that has strong performance guarantees in the *no-regret* model, and (*iii*) an empirical evaluation on two problems of information gathering on the web.

## 2   The assignment learning problem

We consider problems, where we have $K$ positions (e.g., slots for displaying ads), and need to assign to each position an item (e.g., an ad) in order to maximize a utility function (e.g., the revenue from clicks on the ads). We address both the *offline* problem, where the utility function is specified in advance, and the *online* problem, where a sequence of utility functions arrives over time, and we need to repeatedly select a new assignment.

**The Offline Problem.**   In the offline problem we are given sets $P_1, P_2, \ldots, P_K$, where $P_k$ is the set of items that may be placed in position $k$. We assume without loss of generality that these sets are disjoint.[1] An *assignment* is a subset $S \subseteq \mathcal{V}$, where $\mathcal{V} = P_1 \cup P_2 \cup \cdots \cup P_K$ is the set of all items. We call an assignment *feasible*, if at most one item is assigned to each position (i.e., for all $k$, $|S \cap P_k| \leq 1$). We use $\mathcal{P}$ to refer to the set of feasible assignments.

Our goal is to find a feasible assignment maximizing a utility function $f : 2^{\mathcal{V}} \to \mathbb{R}_{\geq 0}$. As we discuss later, many important assignment problems satisfy *submodularity*, a natural diminishing returns property: Assigning a new item to a position $k$ increases the utility more if few elements have been assigned yet, and less if many items have already been assigned. Formally, a utility function $f$ is called submodular, if for all $S \subseteq S'$ and $s \notin S'$ it holds that $f(S \cup \{s\}) - f(S) \geq f(S' \cup \{s\}) - f(S')$. We will also assume $f$ is monotone (i.e., for all $S \subseteq S'$, we have $f(S) \leq f(S')$). Our goal is thus, for a given non-negative, monotone and submodular utility function $f$, to find a feasible assignment $S^*$ of maximum utility, $S^* = \arg\max_{S \in \mathcal{P}} \{f(S)\}$.

This optimization problem is NP-hard. In fact, a stronger negative result holds:

**Theorem 1** ([14]).   *For any $\epsilon > 0$, any algorithm guaranteed to obtain a solution within a factor of $(1 - 1/e + \epsilon)$ of $\max_{S \in \mathcal{P}} \{f(S)\}$ requires exponentially many evaluations of $f$ in the worst case.*

In light of this negative result, we can only hope to obtain a solution that achieves a fraction of $(1 - 1/e)$ of the optimal value. In §3.2 we develop such an algorithm.

**The Online Problem.**   The offline problem is inappropriate to model dynamic settings, where the utility function may change over time, and we need to repeatedly select new assignments, trading off exploration (experimenting with ad display to gain information about the utility function), and exploitation (displaying ads which we believe will maximize utility). More formally, we face a sequential decision problem, where, on each round (which, e.g., corresponds to a user query for a particular term), we want to select an assignment $S_t$ (ads to display). We assume that the sets $P_1$, $P_2, \ldots, P_K$ are fixed in advance for all rounds. After we select the assignment we obtain reward $f_t(S_t)$ for some monotone submodular utility function $f_t$. We call the setting where we do not get any information about $f_t$ beyond the reward the *bandit feedback* model. In contrast, in the *full-information feedback* model we obtain oracle access to $f_t$ (i.e., we can evaluate $f_t$ on arbitrary feasible assignments). Both models arise in real applications, as we show in §5.

The goal is to maximize the total reward we obtain, namely $\sum_t f_t(S_t)$. Following the multiarmed bandit literature, we evaluate our performance after $T$ rounds by comparing our total reward against that obtained by a clairvoyant algorithm with knowledge of the sequence of functions $\langle f_1, \ldots, f_T \rangle$, but with the restriction that it must select the same assignment on each round. The difference between the clairvoyant algorithm's total reward and ours is called our *regret*. The goal is then to develop an algorithm whose expected regret grows sublinearly in the number of rounds; such an algorithm is said to have (or be) *no-regret*. However, since sums of submodular functions remain submodular, the clairvoyant algorithm has to solve an offline assignment problem with $f(S) = \sum_t f_t(S)$. Considering Theorem 1, no polynomial-time algorithm can possibly hope to achieve a no-regret guarantee. To accommodate this fact, we discount the reward of the clairvoyant algorithm by a factor of $(1 - 1/e)$: We define the $(1 - 1/e)$-*regret* of a random sequence $\langle S_1, \ldots, S_T \rangle$ as

$$\left(1 - \frac{1}{e}\right) \cdot \max_{S \in \mathcal{P}} \left\{\sum_{t=1}^T f_t(S)\right\} - \mathbb{E}\left[\sum_{t=1}^T f_t(S_t)\right].$$

---
[1]If the same item can be placed in multiple positions, simply create multiple distinct copies of it.

Our goal is then to develop efficient algorithms whose $(1 - 1/e)$-regret grows sublinearly in $T$.

**Subsumed Models.** Our model generalizes several common models for sponsored search ad selection, and web search results. These include models with *click-through-rates*, in which it is assumed that each (ad, position) pair has some probability $p(a, k)$ of being clicked on, and there is some monetary reward $b(a)$ that is obtained whenever ad $a$ is clicked on. Often, the click-through-rates are assumed to be *separable*, meaning $p(a, k)$ has the functional form $\alpha(a) \cdot \beta(k)$ for some functions $\alpha$ and $\beta$. See [7, 12] for more details on sponsored search ad allocation. Note that in both of these cases, the (expected) reward of a set $S$ of (ad, position) pairs is $\sum_{(a,k) \in S} g(a, k)$ for some nonnegative function $g$. It is easy to verify that such a reward function is monotone submodular. Thus, we can capture this model in our framework by setting $P_k = \mathcal{A} \times \{k\}$, where $\mathcal{A}$ is the set of ads. Another subsumed model, for web search, appears in [16]; it assumes that each user is interested in a particular set of results, and any list of results that intersects this set generates a unit of value; all other lists generate no value (the ordering of results is irrelevant). Again, the reward function is monotone submodular. In this setting, it is desirable to display a diverse set of results in order to maximize the likelihood that at least one of them will interest the user.

Our model is flexible in that we can handle position-dependent effects and diversity considerations simultaneously. For example, we can handle the case that each user $u$ is interested in a particular set $A_u$ of ads and looks at a set $I_u$ of positions, and the reward of an assignment $S$ is any monotone-increasing concave function $g$ of $|S \cap (A_u \times I_u)|$. If $I_u = \{1, 2, \ldots, k\}$ and $g(x) = x$, this models the case where the quality is the number of relevant result that appear in the first $k$ positions. If $I_u$ equals all positions and $g(x) = \min\{x, 1\}$ we recover the model of [16].

## 3 An approximation algorithm for the offline problem

### 3.1 The locally greedy algorithm

A simple approach to the assignment problem is the following greedy procedure: the algorithm steps through all $K$ positions (according to some fixed, arbitrary ordering). For position $k$, it simply chooses the item that increases the total value as much as possible, i.e., it chooses

$$s_k = \arg\max_{s \in P_k} \{f(\{s_1, \ldots, s_{k-1}\} + s)\},$$

where, for a set $S$ and element $e$, we write $S + e$ for $S \cup \{e\}$. Perhaps surprisingly, no matter which ordering over the positions is chosen, this so-called *locally greedy algorithm* produces an assignment that obtains at least half the optimal value [8]. In fact, the following more general result holds. We will use this lemma in the analysis of our improved offline algorithm, which uses the locally greedy algorithm as a subroutine.

**Lemma 2.** *Suppose $f : 2^\mathcal{V} \to \mathbb{R}_{\geq 0}$ is of the form $f(S) = f_0(S) + \sum_{k=1}^K f_k(S \cap P_k)$ where $f_0 : 2^\mathcal{V} \to \mathbb{R}_{\geq 0}$ is monotone submodular, and $f_k : P_k \to \mathbb{R}_{\geq 0}$ is arbitrary for $k \geq 1$. Let $L$ be the solution returned by the locally greedy algorithm. Then $f(L) + f_0(L) \geq \max_{S \in \mathcal{P}} \{f(S)\}$.*

The proof is given in Appendix A. Observe that in the special case where $f_k = 0$ for all $k \geq 1$, Lemma 2 says that $f(L) \geq \frac{1}{2} \max_{S \in \mathcal{P}} f(S)$.

The following example shows that the $1/2$ approximation ratio is tight. Consider an instance of the ad allocation problem with two ads, two positions and two users, Alice and Bob. Alice is interested in ad 1, but has a very short attention span: She will only click on the ad if it appears in the first position. Bob is interested in ad 2, and will look through all positions. Now suppose that Alice searches slightly less frequently (with probability $\frac{1}{2} - \varepsilon$) than Bob (who searches with probability $\frac{1}{2} + \varepsilon$). The greedy algorithm first chooses the ad to assign to slot 1. Since the ad is more likely to be shown to Bob, the algorithm chooses ad 2, with an expected utility of $\frac{1}{2} + \varepsilon$. Since Alice will only look at position 1, no ad assigned to slot 2 can increase the expected utility further. On the other hand, the optimal solution is to assign ad 1 to slot 1, and ad 2 to slot 2, with an expected utility of 1.

### 3.2 An algorithm with optimal approximation ratio

We now present an algorithm that achieves the optimal approximation ratio of $1 - 1/e$, improving on the $\frac{1}{2}$ approximation for the locally greedy algorithm. Our algorithm associates with each partition $P_k$ a *color* $c_k$ from a palette $[C]$ of $C$ colors, where we use the notation $[n] = \{1, 2, \ldots, n\}$. For any set $S \subseteq \mathcal{V} \times [C]$ and vector $\vec{c} = (c_1, \ldots, c_K)$, define

$$\mathsf{sample}_{\vec{c}}(S) = \bigcup_{k=1}^{K} \{x \in P_k : (x, c_k) \in S\} \ .$$

Given a set $S$ of (item, color) pairs, which we may think of as labeling each item with one or more colors, $\mathsf{sample}_{\vec{c}}(S)$ returns a set containing each item $x$ that is labeled with whatever color $\vec{c}$ assigns to the partition that contains $x$. Let $F(S)$ denote the expected value of $f(\mathsf{sample}_{\vec{c}}(S))$ when each color $c_k$ is selected uniformly at random from $[C]$. We consider the following algorithm.

---

**Algorithm:** TABULARGREEDY

**Input**: integer $C$, sets $P_1, P_2, \ldots, P_K$, function $f : 2^{\mathcal{V}} \to \mathbb{R}_{\geq 0}$ (where $\mathcal{V} = \bigcup_{k=1}^{K} P_k$)

**set** $G := \emptyset$.
**for** $c$ **from** 1 **to** $C$ **do**                                                  /* For each color */
    **for** $k$ **from** 1 **to** $K$ **do**                                    /* For each partition */
        **set** $g_{k,c} = \arg\max_{x \in P_k \times \{c\}} \{F(G + x)\}$       /* Greedily pick $g_{k,c}$ */
        **set** $G := G + g_{k,c}$;
**for each** $k \in [K]$, choose $c_k$ uniformly at random from $[C]$.
**return** $\mathsf{sample}_{\vec{c}}(G)$, where $\vec{c} := (c_1, \ldots, c_K)$.

---

Observe that when $C = 1$, the $\mathsf{sample}_{\vec{c}}$ function is deterministic and TABULARGREEDY is simply the locally greedy algorithm from §3.1. In the limit as $C \to \infty$, the TABULARGREEDY can intuitively be viewed as an algorithm for a continuous extension of the problem followed by a rounding procedure, in the same spirit as Vondrák's *continuous-greedy algorithm* [4]. In our case, the continuous extension is to compute a probability distribution $D_k$ for each position $k$ with support in $P_k$ (plus a special "select nothing" outcome), such that if we independently sample an element $x_k$ from $D_k$, $\mathbb{E}\left[f(\{x_1, \ldots, x_K\})\right]$ is maximized. It turns out that if the positions individually, greedily, and in round-robin fashion, add infinitesimal units of probability mass to their distributions so as to maximize this objective function, they achieve the same objective function value as if, rather than making decisions in a round-robin fashion, they had cooperated and added the combination of $K$ infinitesimal probability mass units (one per position) that greedily maximizes the objective function. The latter process, in turn, can be shown to be equivalent to a greedy algorithm for maximizing a (different) submodular function subject to a cardinality constraint, which implies that it achieves a $1 - 1/e$ approximation ratio [15]. TABULARGREEDY represents a tradeoff between these two extremes; its performance is summarized by the following theorem. For now, we assume that the $\arg\max$ in the inner loop is computed exactly. In Appendix A we bound the performance loss that results from approximating the $\arg\max$ (e.g., by estimating $F$ by repeated sampling).

**Theorem 3.** *Suppose $f$ is monotone submodular. Then $F(G) \geq \beta(K, C) \cdot \max_{S \in \mathcal{P}} \{f(S)\}$, where $\beta(K, C)$ is defined as $1 - (1 - \frac{1}{C})^C - \binom{K}{2} C^{-1}$.*

It follows that, for any $\varepsilon > 0$, TABULARGREEDY achieves a $(1 - 1/e - \varepsilon)$ approximation factor using a number of colors that is polynomial in $K$ and $1/\varepsilon$. The theorem will follow immediately from the combination of two key lemmas, which we now prove. Informally, Lemma 4 analyzes the approximation error due to the outer greedy loop of the algorithm, while Lemma 5 analyzes the approximation error due to the inner loop.

**Lemma 4.** *Let $G_c = \{g_{1,c}, g_{2,c}, \ldots, g_{K,c}\}$, and let $G_c^- = G_1 \cup G_2 \cup \ldots \cup G_{c-1}$. For each color $c$, choose $E_c \in \mathbb{R}$ such that $F(G_c^- \cup G_c) \geq \max_{x \in \mathcal{R}_c} \{F(G_c^- \cup x)\} - E_c$ where $\mathcal{R}_c := \{R : \forall k \in [K], |R \cap (P_k \times \{c\})| = 1\}$ is the set of all possible choices for $G_c$. Then*

$$F(G) \geq \beta(C) \cdot \max_{S \in \mathcal{P}} \{f(S)\} - \sum_{c=1}^C E_c \,. \tag{3.1}$$

*where $\beta(C) = 1 - \left(1 - \frac{1}{C}\right)^C$.*

*Proof (Sketch).* We will refer to an element $R$ of $\mathcal{R}_c$ as a *row*, and to $c$ as the color of the row. Let $\mathcal{R}_{[C]} := \bigcup_{c=1}^C \mathcal{R}_c$ be the set of all rows. Consider the function $H : 2^{\mathcal{R}_{[C]}} \to \mathbb{R}_{\geq 0}$, defined as $H(\mathcal{R}) = F\left(\bigcup_{R \in \mathcal{R}} R\right)$. We will prove the lemma in three steps: (*i*) $H$ is monotone submodular, (*ii*) TABULARGREEDY is simply the locally greedy algorithm for finding a set of $K$ rows that maximizes $H$, where the $c^{\text{th}}$ greedy step is performed with additive error $E_c$, and (*iii*) TABULARGREEDY obtains the guarantee (3.1) for maximizing $H$, and this implies the same ratio for maximizing $F$.

To show that $H$ is monotone submodular, it suffices to show that $F$ is monotone submodular. Because $F(S) = \mathbb{E}_{\vec{c}}[f(\mathsf{sample}_{\vec{c}}(S))]$, and because a convex combination of monotone submodular functions is monotone submodular, it suffices to show that for any particular coloring $\vec{c}$, the function $f(\mathsf{sample}_{\vec{c}}(S))$ is monotone submodular. This follows from the definition of $\mathsf{sample}$ and the fact that $f$ is monotone submodular.

The second claim is true by inspection. To prove the third claim, we use the fact that $F(G_c^- \cup R)$ can always be maximized by choosing a row $R \in \mathcal{R}_c$. Informally, this is because $F(G_c^- \cup R)$ can always be maximized by choosing a row whose color has not already been used, and all colors $\geq c$ are interchangeable. For problems with this special property, it is known that the locally greedy algorithm obtains an approximation ratio of $\beta(C) = 1 - (1 - \frac{1}{C})^C$ [15]. Theorem 6 of [17] extends this result to handle additive error, and yields

$$F(G) = H(\{G_1, G_2, \ldots, G_C\}) \geq \beta(C) \cdot \max_{\mathcal{R} \subseteq \mathcal{R}_{[C]} : |\mathcal{R}| \leq C} \{H(\mathcal{R})\} - \sum_{c=1}^C E_c \,.$$

To complete the proof, it suffices to show that $\max_{\mathcal{R} \subseteq \mathcal{R}_{[C]} : |\mathcal{R}| \leq C} \{H(\mathcal{R})\} \geq \max_{S \in \mathcal{P}} \{f(S)\}$. This follows from the fact that for any assignment $S \in \mathcal{P}$, we can find a set $\mathcal{R}(S)$ of $C$ rows such that $\mathsf{sample}_{\vec{c}}(\bigcup_{R \in \mathcal{R}(S)} R) = S$ with probability 1, and therefore $H(\mathcal{R}(S)) = f(S)$. □

We now bound the performance of the the inner loop of TABULARGREEDY.

**Lemma 5.** *Let $f^* = \max_{S \in \mathcal{P}} \{f(S)\}$, and let $G_c$, $G_c^-$, and $\mathcal{R}_c$ be defined as in the statement of Lemma 4. Then, for any $c \in [C]$,*

$$F(G_c^- \cup G_c) \geq \max_{R \in \mathcal{R}_c} \{F(G_c^- \cup R)\} - \binom{K}{2} C^{-2} f^* \,.$$

*Proof (Sketch).* Let $N$ denote the number of partitions whose color is $c$. For $R \in \mathcal{R}_c$, let $\Delta_{\vec{c}}(R) := f(\mathsf{sample}_{\vec{c}}(G_c^- \cup R)) - f(\mathsf{sample}_{\vec{c}}(G_c^-))$, and let $F_c(R) := F(G_c^- \cup R) - F(G_c^-)$. By definition, $F_c(R) = \mathbb{E}_{\vec{c}}[\Delta_{\vec{c}}(R)] = \mathbb{P}[N = 1] \mathbb{E}_{\vec{c}}[\Delta_{\vec{c}}(R)|N = 1] + \mathbb{P}[N \geq 2] \mathbb{E}_{\vec{c}}[\Delta_{\vec{c}}(R)|N \geq 2]$, where we have used the fact that $\Delta_{\vec{c}}(R) = 0$ when $N = 0$. The idea of the proof is that the first of these terms dominates as $C \to \infty$, and that $\mathbb{E}_{\vec{c}}[\Delta_{\vec{c}}(R)|N = 1]$ can be optimized exactly simply by optimizing each element of $P_k \times \{c\}$ independently. Specifically, it can be seen that $\mathbb{E}_{\vec{c}}[\Delta_{\vec{c}}(R)|N = 1] = \sum_{k=1}^K f_k(R \cap (P_k \times \{c\}))$ for suitable $f_k$. Additionally, $f_0(R) = \mathbb{P}[N \geq 2] \mathbb{E}_{\vec{c}}[\Delta_{\vec{c}}(R)|N \geq 2]$ is a monotone submodular function of a set of (item, color) pairs, for the same reasons $F$ is. Applying Lemma 2 with these $\{f_k : k \geq 0\}$ yields

$$F_c(G_c) + \mathbb{P}[N \geq 2] \mathbb{E}_{\vec{c}}[\Delta_{\vec{c}}(G_c)|N \geq 2] \geq \max_{R \in \mathcal{R}_c} \{F_c(R)\} \,.$$

To complete the proof, it suffices to show that $\mathbb{P}[N \geq 2] \leq \binom{K}{2} C^{-2}$ and that $\mathbb{E}_{\vec{c}}[\Delta_{\vec{c}}(G_c)|N \geq 2] \leq f^*$. The first inequality holds because, if we let $M$ be the number of *pairs* of partitions that are both assigned color $c$, we have $\mathbb{P}[N \geq 2] = \mathbb{P}[M \geq 1] \leq \mathbb{E}[M] = \binom{K}{2} C^{-2}$. The second inequality follows from the fact that for any $\vec{c}$ we have $\Delta_{\vec{c}}(G_c) \leq f(\mathsf{sample}_{\vec{c}}(G_c^- \cup G_c)) \leq f^*$. □

# 4 An algorithm for online learning of assignments

We now transform the offline algorithm of §3.2 into an online algorithm. The high-level idea behind this transformation is to replace each greedy decision made by the offline algorithm with a no-regret online algorithm. A similar approach was used in [16] and [18] to obtain an online algorithm for different (simpler) online problems.

---

**Algorithm:** TGBANDIT (described in the full-information feedback model)
**Input**: sets $P_1, P_2, \ldots, P_K$

**for each** $k \in [K]$ and $c \in [C]$, let $\mathcal{E}_{k,c}$ be a no-regret algorithm with action set $P_k \times \{c\}$.
**for** $t$ **from** $1$ **to** $T$ **do**
    **for each** $k \in [K]$ and $c \in [C]$, let $g_{k,c}^t \in P_k \times \{c\}$ be the action selected by $\mathcal{E}_{k,c}$
    **for each** $k \in [K]$, choose $c_k$ uniformly at random from $[C]$. Define $\vec{c} = (c_1, \ldots, c_K)$.
    select the set $G_t = \mathsf{sample}_{\vec{c}}\left(\left\{g_{k,c}^t : k \in [K], c \in [C]\right\}\right)$
    observe $f_t$, and let $\bar{F}_t(S) := f_t(\mathsf{sample}_{\vec{c}}(S))$
    **for each** $k \in [K]$, $c \in [C]$ **do**
        define $G_{k,c}^{t-} \equiv \left\{g_{k',c'}^t : k' \in [K], c' < c\right\} \cup \left\{g_{k',c}^t : k' < k\right\}$
        **for each** $x \in P_k \times \{c\}$, feed back $\bar{F}_t(G_{k,c}^{t-} + x)$ to $\mathcal{E}_{k,c}$ as the reward for choosing $x$

---

The following theorem summarizes the performance of TGBANDIT.

**Theorem 6.** *Let $r_{k,c}$ be the regret of $\mathcal{E}_{k,c}$, and let $\beta(K,C) = 1 - \left(1 - \frac{1}{C}\right)^C - \binom{K}{2}C^{-1}$. Then*

$$\mathbb{E}\left[\sum_{t=1}^T f_t(G_t)\right] \geq \beta(K,C) \cdot \max_{S \in \mathcal{P}}\left\{\sum_{t=1}^T f_t(S)\right\} - \mathbb{E}\left[\sum_{k=1}^K \sum_{c=1}^C r_{k,c}\right].$$

Observe that Theorem 6 is similar to Theorem 3, with the addition of the $\mathbb{E}[r_{k,c}]$ terms. The idea of the proof is to view TGBANDIT as a version of TABULARGREEDY that, instead of greedily selecting single (element,color) pairs $g_{k,c} \in P_k \times \{c\}$, greedily selects (element vector, color) pairs $\vec{g}_{k,c} \in P_k^T \times \{c\}$ (this greedy decision is made with additive error $r_{k,c}$, which is the source of the extra terms). Once this correspondence is established, the theorem follows along the lines of Theorem 3. For a proof, see Appendix A.

**Corollary 7.** *If TGBANDIT is run with randomized weighted majority [5] as the subroutine, then*

$$\mathbb{E}\left[\sum_{t=1}^T f_t(G_t)\right] \geq \beta(K,C) \cdot \max_{S \in \mathcal{P}}\left\{\sum_{t=1}^T f_t(S)\right\} - O\left(C \sum_{k=1}^K \sqrt{T \log |P_k|}\right).$$

*where $\beta(K,C) = 1 - \left(1 - \frac{1}{C}\right)^C - \binom{K}{2}C^{-1}$.*

Optimizing for $C$ in Corollary 7 yields $(1 - \frac{1}{e})$-regret $\tilde{\Theta}(K^{3/2}T^{1/4}\sqrt{\mathsf{OPT}})$ ignoring logarithmic factors, where $\mathsf{OPT} := \max_{S \in \mathcal{P}}\left\{\sum_{t=1}^T f_t(S)\right\}$ is the value of the static optimum.

**Dealing with bandit feedback.** TGBANDIT can be modified to work in the bandit feedback model. The idea behind this modification is that on each round we "explore" with some small probability, in such a way that on each round we obtain an unbiased estimate of the desired feedback values $\bar{F}_t(G_{k,c}^{t-} + x)$ for each $k \in [K], c \in [C]$, and $x \in P_k$. This technique can be used to achieve a bound similar to the one stated in Corollary 7, but with an additive regret term of $O\left((T|\mathcal{V}|CK)^{\frac{2}{3}}(\log |\mathcal{V}|)^{\frac{1}{3}}\right)$.

**Stronger notions of regret.** By substituting in different algorithms for the subroutines $\mathcal{E}_{k,c}$, we can obtain additional guarantees. For example, Blum and Mansour [3] consider online problems in which we are given *time-selection functions* $I_1, I_2, \ldots, I_M$. Each time-selection function $I : [T] \to [0,1]$ associates a weight with each round, and defines a corresponding weighted notion of regret in the natural way. Blum and Mansour's algorithm guarantees low weighted regret with respect to all $M$ time selection functions simultaneously. This can be used to obtain low regret with respect to different (possibly overlapping) windows of time simultaneously, or to obtain low regret with respect to subsets of rounds that have particular features. By using their algorithm as a subroutine within TGBANDIT, we get similar guarantees, both in the full information and bandit feedback models.

## 5 Evaluation

We evaluate TGBANDIT experimentally on two applications: Learning to rank blogs that are effective in detecting cascades of information, and allocating advertisements to maximize revenue.

### 5.1 Online learning of diverse blog rankings

We consider the problem of ranking a set of blogs and news sources on the web. Our approach is based on the following idea: A blogger writes a posting, and, after some time, other postings link to it, forming cascades of information propagating through the network of blogs.

More formally, an information cascade is a directed acyclic graph of vertices (each vertex corresponds to a posting at some blog), where edges are annotated by the time difference between the postings. Based on this notion of an information cascade, we would like to select blogs that detect big cascades (containing many nodes) as early as possible (i.e., we want to learn about an important event before most other readers). In [13] it is shown how one can formalize this notion of utility using a monotone submodular function that measures the informativeness of a subset of blogs. Optimizing the submodular function yields a small set of blogs that "covers" most cascades. This utility function prefers *diverse* sets of blogs, minimizing the overlap of the detected cascades, and therefore minimizing redundancy.

The work by [13] leaves two major shortcomings: Firstly, they select a *set* of blogs rather than a *ranking*, which is of practical importance for the presentation on a web service. Secondly, they do not address the problem of *sequential* prediction, where the set of blogs must be updated dynamically over time. In this paper, we address these shortcomings.

**Results on offline blog ranking.** In order to model the blog ranking problem, we adopt the assumption that different users have different attention spans: Each user will only consider blogs appearing in a particular subset of positions. In our experiments, we assume that the probability that a user is willing to look at position $k$ is proportional to $\gamma^k$, for some discount factor $0 < \gamma < 1$. More formally, let $g$ be the monotone submodular function measuring the informativeness of any set of blogs, defined as in [13]. Let $P_k = \mathcal{B} \times \{k\}$, where $\mathcal{B}$ is the set of blogs. Given an assignment $S \in \mathcal{P}$, let $S^{[k]} = S \cap \{P_1 \cup P_2 \cup \ldots \cup P_k\}$ be the assignment of blogs to positions 1 through $k$. We define the *discounted* value of the assignment $S$ as $f(S) = \sum_{k=1}^{K} \gamma^k \left( g(S^{[k]}) - g(S^{[k-1]}) \right)$. It can be seen that $f : 2^{\mathcal{V}} \to \mathbb{R}_{\geq 0}$ is monotone submodular.

For our experiments, we use the data set of [13], consisting of 45,192 blogs, 16,551 cascades, and 2 million postings collected during 12 months of 2006. We use the *population affected* objective of [13], and use a discount factor of $\gamma = 0.8$. Based on this data, we run our TABULARGREEDY algorithm with varying numbers of colors $C$ on the blog data set. Fig. 1(a) presents the results of this experiment. For each value of $C$, we generate 200 rankings, and report both the average performance and the maximum performance over the 200 trials. Increasing $C$ leads to an improved performance over the locally greedy algorithm ($C = 1$).

**Results on online learning of blog rankings.** We now consider the online problem where on each round $t$ we want to output a ranking $S_t$. After we select the ranking, a new set of cascades occurs, modeled using a separate submodular function $f_t$, and we obtain a reward of $f_t(S_t)$. In our experiments, we choose one assignment per day, and define $f_t$ as the utility associated with the cascades occurring on that day. Note that $f_t$ allows us to evaluate the performance of any possible ranking $S_t$, hence we can apply TGBANDIT in the *full-information feedback* model.

We compare the performance of our online algorithm using $C = 1$ and $C = 4$. Fig. 1(b) presents the average cumulative reward gained over time by both algorithms. We normalize the average reward by the utility achieved by the TABULARGREEDY algorithm (with $C = 1$) applied to the entire data set.

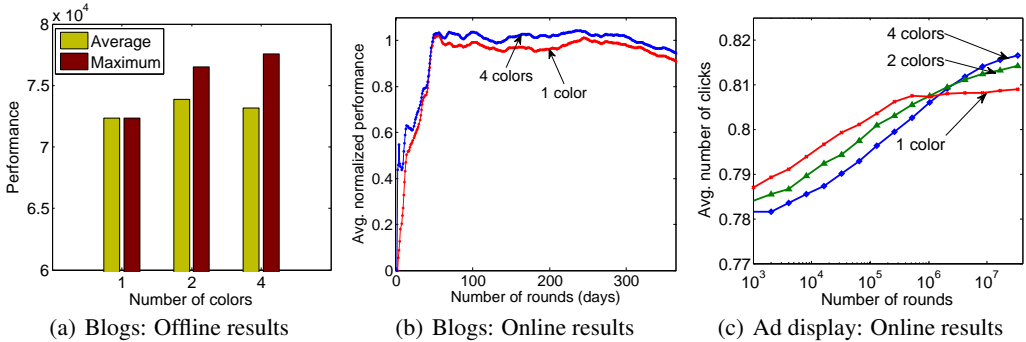

Figure 1: (a,b) Results for discounted blog ranking ($\gamma = 0.8$), in offline (a) and online (b) setting. (c) Performance of TGBANDIT with $C = 1, 2,$ and $4$ colors for the sponsored search ad selection problem (each round is a query). Note that $C = 1$ corresponds to the online algorithm of [16, 18].

Fig. 1(b) shows that the performance of both algorithms rapidly (within the first 47 rounds) converges to the performance of the offline algorithm. The TGBANDIT algorithm with $C = 4$ levels out at an approximately 4% higher reward than the algorithm with $C = 1$.

### 5.2 Online ad display

We evaluate TGBANDIT for the sponsored search ad selection problem in a simple Markovian model incorporating the value of diverse results and complex position-dependence among clicks. In this model, each user $u$ is defined by two sets of probabilities: $p_{\text{click}}(a)$ for each ad $a \in \mathcal{A}$, and $p_{\text{abandon}}(k)$ for each position $k \in [K]$. When presented an assignment of ads $\{a_1, a_2, \ldots, a_K\}$, where $a_k$ occupies position $k$, the user scans the positions in increasing order. For each position $k$, the user clicks on $a_k$ with probability $p_{\text{click}}(a_k)$, leaving the results page forever. Otherwise, with probability $(1 - p_{\text{click}}(a_k)) \cdot p_{\text{abandon}}(k)$, the user loses interest and abandons the results without clicking on anything. Finally, with probability $(1 - p_{\text{click}}(a_k)) \cdot (1 - p_{\text{abandon}}(k))$, the user proceeds to look at position $k + 1$. The reward function $f_t$ is the number of clicks, which is either zero or one. We only receive information about $f_t(S_t)$ (i.e., *bandit feedback*).

In our evaluation, there are two types of users: those interested in all positions ($p_{\text{abandon}} \equiv 0$), and those that quickly lose interest ($p_{\text{abandon}} \equiv 0.5$). For both types of users we select $p_{\text{click}}(a)$ uniformly at random from $[0, 1]$, independently for each ad $a$ (once chosen, $p_{\text{click}}$ is fixed for all rounds). We use $K = 6$ positions, and 6 available ads. In Fig. 1(c) we compare the performance of TGBANDIT with $C = 4$ to the online algorithm of [16, 18], based on the average of 100 experiments. The latter algorithm is equivalent to running TGBANDIT with $C = 1$. The former achieves parity with the latter after roughly $10^6$ rounds, and dominates thereafter.

It can be shown that with several different types of users with distinct $p_{\text{click}}(\cdot)$ functions the offline problem of finding an assignment within $1 - \frac{1}{e} + \varepsilon$ of optimal is NP-hard. This is in contrast to the case in which $p_{\text{click}}$ and $p_{\text{abandon}}$ are the *same* for all users; in this case the offline problem simply requires finding an optimal policy for a Markov decision process, which can be done efficiently using well-known algorithms. A slightly different Markov model of user behavior which is efficiently solvable was considered in [1]. In that model, $p_{\text{click}}$ and $p_{\text{abandon}}$ are the same for all users, and $p_{\text{abandon}}$ is a function of the ad in the slot currently being scanned rather than its index.

## 6 Related Work

For a general introduction to the literature on submodular function maximization, see [19]. For applications of submodularity to machine learning and AI see [10, 11].

Our offline problem is known as *maximizing a monotone submodular function subject to a (simple) partition matroid constraint* in the operations research and theoretical computer science communities. The study of this problem culminated in the elegant $(1-1/e)$ approximation algorithm of Vondrák [20] and a matching unconditional lower bound of Mirrokni *et al.* [14]. Vondrák's algorithm, called the *continuous-greedy algorithm*, has also been extended to handle arbitrary matroid constraints [4]. The continuous-greedy algorithm, however, cannot be applied to our problem directly, because it requires the ability to sample $f(\cdot)$ on *infeasible* sets $S \notin \mathcal{P}$. In our context, this means it must have the ability to ask (for example) what the revenue will be if ads $a_1$ and $a_2$ are placed in position #1 *simultaneously*. We do not know how to answer such questions in a way that leads to meaningful performance guarantees.

In the online setting, the most closely related work is that of Streeter and Golovin [18]. Like us, they consider sequences of monotone submodular reward functions that arrive online, and develop an online algorithm that uses multi-armed bandit algorithms as subroutines. The key difference from our work is that, as in [16], they are concerned with selecting a *set* of $K$ items rather than the more general problem of selecting an *assignment* of items to positions addressed in this paper. Kakade *et al.* [9] considered the general problem of using $\alpha$-approximation algorithms to construct no $\alpha$-regret online algorithms, and essentially proved it could be done for the class of linear optimization problems in which the cost function has the form $c(S, w)$ for a solution $S$ and weight vector $w$, and $c(S, w)$ is linear in $w$. However, their result is orthogonal to ours, because our objective function is submodular and not linear [2].

## 7 Conclusions

In this paper, we showed that important problems, such as ad display in sponsored search and computing diverse rankings of information sources on the web, require optimizing assignments under submodular utility functions. We developed an efficient algorithm, TABULARGREEDY, which obtains the optimal approximation ratio of $(1 - 1/e)$ for this NP-hard optimization problem. We also developed an online algorithm, TGBANDIT, that asymptotically achieves no $(1 - 1/e)$-regret for the problem of repeatedly selecting informative assignments, under the full-information and bandit-feedback settings. Finally, we demonstrated that our algorithm outperforms previous work on two real world problems, namely online ranking of informative blogs and ad allocation.

## Acknowledgments

This work was supported in part by Microsoft Corporation through a gift as well as through the Center for Computational Thinking at Carnegie Mellon, by NSF ITR grant CCR-0122581 (The Aladdin Center), and by ONR grant N00014-09-1-1044.

## References

[1] Gagan Aggarwal, Jon Feldman, S. Muthukrishnan, and Martin Pál. Sponsored search auctions with markovian users. In *WINE '08: Proceedings of the 4th International Workshop on Internet and Network Economics*, pages 621–628, Berlin, Heidelberg, 2008. Springer-Verlag.

[2] Peter Auer, Nicolò Cesa-Bianchi, Yoav Freund, and Robert E. Schapire. The nonstochastic multiarmed bandit problem. *SIAM Journal on Computing*, 32(1):48–77, 2002.

[3] Avrim Blum and Yishay Mansour. From external to internal regret. *Journal of Machine Learning Research*, 8:1307–1324, 2007.

[4] Gruia Calinescu, Chandra Chekuri, Martin Pál, and Jan Vondrák. Maximizing a submodular set function subject to a matroid constraint. *SIAM Journal on Computing*. To appear.

[5] Nicolò Cesa-Bianchi, Yoav Freund, David Haussler, David P. Helmbold, Robert E. Schapire, and Manfred K. Warmuth. How to use expert advice. *J. ACM*, 44(3):427–485, 1997.

[6] Benjamin Edelman, Michael Ostrovsky, and Michael Schwarz. Internet advertising and the generalized second price auction: Selling billions of dollars worth of keywords. *American Economic Review*, 97(1):242–259, 2007.

[7] Jon Feldman and S. Muthukrishnan. Algorithmic methods for sponsored search advertising. In Zhen Liu and Cathy H. Xia, editors, *Performance Modeling and Engineering*. 2008.

---

[2]Of course, it is possible to linearize a submodular function by using a separate dimension for every possible function argument, but this results in an exponential number of dimensions, which leads to exponentially worse convergence time and regret bounds for the algorithms in [9] relative to TGBANDIT.

[8] Marshall L. Fisher, George L. Nemhauser, and Laurence A. Wolsey. An analysis of approximations for maximizing submodular set functions - II. *Mathematical Programming Study*, (8):73–87, 1978.

[9] Sham M. Kakade, Adam Tauman Kalai, and Katrina Ligett. Playing games with approximation algorithms. In *STOC '07: Proceedings of the thirty-ninth annual ACM symposium on Theory of computing*, pages 546–555, New York, NY, USA, 2007. ACM.

[10] Andreas Krause and Carlos Guestrin. Beyond convexity: Submodularity in machine learning. Tutorial at ICML 2008. http://www.select.cs.cmu.edu/tutorials/icml08submodularity.html.

[11] Andreas Krause and Carlos Guestrin. Near-optimal observation selection using submodular functions. In *AAAI Nectar track*, 2007.

[12] Sébastien Lahaie, David M. Pennock, Amin Saberi, and Rakesh V. Vohra. Sponsored search auctions. In Noam Nisan, Tim Roughgarden, Eva Tardos, and Vijay V. Vazirani, editors, *Algorithmic Game Theory*. Cambridge University Press, New York, NY, USA, 2007.

[13] Jure Leskovec, Andreas Krause, Carlos Guestrin, Christos Faloutsos, Jeanne VanBriesen, and Natalie Glance. Cost-effective outbreak detection in networks. In *ACM SIGKDD International Conference on Knowledge Discovery and Data Mining (KDD)*, August 2007.

[14] Vahab Mirrokni, Michael Schapira, and Jan Vondrák. Tight information-theoretic lower bounds for welfare maximization in combinatorial auctions. In *EC '08: Proceedings of the 9th ACM conference on Electronic commerce*, pages 70–77, New York, NY, USA, 2008. ACM.

[15] George L. Nemhauser, Laurence A. Wolsey, and Marshall L. Fisher. An analysis of approximations for maximizing submodular set functions - I. *Mathematical Programming*, 14(1):265–294, 1978.

[16] Filip Radlinski, Robert Kleinberg, and Thorsten Joachims. Learning diverse rankings with multi-armed bandits. In *ICML*, pages 784–791, 2008.

[17] Matthew Streeter and Daniel Golovin. An online algorithm for maximizing submodular functions. Technical Report CMU-CS-07-171, Carnegie Mellon University, 2007.

[18] Matthew Streeter and Daniel Golovin. An online algorithm for maximizing submodular functions. In *NIPS*, pages 1577–1584, 2008.

[19] Jan Vondrák. *Submodularity in Combinatorial Optimization*. PhD thesis, Charles University, Prague, Czech Republic, 2007.

[20] Jan Vondrák. Optimal approximation for the submodular welfare problem in the value oracle model. In *STOC '08: Proceedings of the 40th annual ACM symposium on Theory of computing*, pages 67–74, New York, NY, USA, 2008. ACM.

## Appendix A: Proofs

Lemma 2 is a corollary of the following more general lemma. The difference between the two lemmas is that, unlike Lemma 2, Lemma 8 allows for the possibility that each arg max in the locally greedy algorithm is evaluated with additive error. We will need this result in analyzing TGBANDIT later on.

**Lemma 8.** *Let $f : \mathcal{P} \to \mathbb{R}_{\geq 0}$ be a function of the form $f(S) = f_0(S) + \sum_{k=1}^{K} f_k(S \cap P_k)$, where $f_0 : 2^{\mathcal{V}} \to \mathbb{R}_{\geq 0}$ is monotone submodular, and $f_k : P_k \to \mathbb{R}_{\geq 0}$ is arbitrary for $k \geq 1$. Let $L = \{\ell_1, \ell_2, \ldots, \ell_K\}$, where $\ell_k \in P_k$ (for $1 \leq k \leq K$). Suppose that for any $k$,*

$$f(\{\ell_1, \ell_2, \ldots, \ell_k\}) \geq \max_{x \in P_k} \{f(\{\ell_1, \ell_2, \ldots, \ell_{k-1}\} + x)\} - \varepsilon_k \,. \tag{7.1}$$

*Then*

$$f(L) + f_0(L) \geq \max_{S \in \mathcal{P}} \{f(S)\} - \sum_{k=1}^{K} \varepsilon_k \,.$$

*Proof.* Let $\mathsf{OPT} = \arg\max_{S \in \mathcal{P}} \{f(S)\}$, and let $\mathsf{OPT} = \{o_1, o_2, \ldots, o_K\}$, where $o_k \in P_k$ (for $1 \leq k \leq K$). Define

$$\Delta_k(L) := f_k(\{o_k\}) + f_0(L \cup \{o_1, o_2, \ldots, o_k\}) - f_0(L \cup \{o_1, o_2, \ldots, o_{k-1}\}) \,.$$

Let $L_k = \{\ell_1, \ell_2, \ldots, \ell_{k-1}\}$. Using submodularity of $f_0$, we have
$$\Delta_k(L) \leq f_k(\{o_k\}) + f_0(L_k + o_k) - f_0(L_k)$$
$$= f(L_k + o_k) - f(L_k)$$
$$\leq f(L_k + \ell_k) - f(L_k) + \varepsilon_k .$$

Then, using monotonicity of $f_0$,
$$f(\mathsf{OPT}) \leq f_0(L \cup \mathsf{OPT}) + \sum_{k=1}^{K} f_k(\mathsf{OPT} \cap P_k)$$
$$= f_0(L) + \sum_{k=1}^{K} \Delta_k(L)$$
$$\leq f_0(L) + \sum_{k=1}^{K} f(L_k + \ell_k) - f(L_k) + \varepsilon_k$$
$$= f_0(L) + f(L) - f(\emptyset) + \sum_{k=1}^{K} \varepsilon_k .$$

Rearranging this inequality and using $f(\emptyset) \geq 0$ completes the proof. □

To analyze TGBANDIT, we will also need Theorem 9, which is a generalization of Theorem 3.

**Theorem 9.** *Suppose $f$ is monotone submodular. Let $G = \{g_{k,c} : k \in [K], c \in [C]\}$, where $g_{k,c} \in P_k$ for all $k$ and $c$. Suppose that for all $k \in [K]$ and $c \in [C]$,*
$$F(G_{k,c}^- + g_{k,c}) \geq \max_{x \in P_k \times \{c\}} \left\{ G_{k,c}^- + x \right\} - \varepsilon_{k,c} \tag{7.2}$$
*where $G_{k,c}^- = \{g_{k',c'} : k' \in [K], c' < c\} \cup \{g_{k',c} : k' < k\}$ (i.e., $G_{k,c}^-$ equals $G$ just before $g_{k,c}$ is added). Then $f(G) \geq \beta(C,K) \cdot \max_{S \in \mathcal{P}} \{f(S)\} - \sum_{k=1}^{K} \sum_{c=1}^{C} \varepsilon_{k,c}$, where $\beta(C,K)$ is defined as $1 - (1 - \frac{1}{C})^C - \binom{K}{2} C^{-1}$.*

*Proof.* The proof is identical to the proof of Theorem 3 in the main text, using Lemma 8 in place of Lemma 2. □

Finally, we prove Theorem 6, which we restate here for convenience.

*Theorem 6* 1. *Let $r_{k,c}$ be the regret of $\mathcal{E}_{k,c}$, and let $\beta(K,C) = 1 - \left(1 - \frac{1}{C}\right)^C - \binom{K}{2} C^{-1}$. Then*
$$\mathbb{E}\left[\sum_{t=1}^{T} f_t(G_t)\right] \geq \beta(K,C) \cdot \max_{S \in \mathcal{P}} \left\{ \sum_{t=1}^{T} f_t(S) \right\} - \mathbb{E}\left[\sum_{k=1}^{K} \sum_{c=1}^{C} r_{k,c}\right] .$$

*Proof.* The idea of the proof is to view TGBANDIT as a version of TABULARGREEDY that, instead of greedily selecting single (element,color) pairs $g_{k,c} \in P_k \times \{c\}$, greedily selects (element vector, color) pairs $\vec{g}_{k,c} \in P_k^T \times \{c\}$, where $T$ is the number of rounds.

First note that for any $k \in [K]$, $c \in [C]$, and $x \in P_k \times \{c\}$, by definition of $r_{k,c}$ we have
$$\sum_{t=1}^{T} \bar{F}_t \left( G_{k,c}^t + g_{k,c}^t \right) \geq \left( \sum_{t=1}^{T} \bar{F}_t \left( G_{k,c}^t + x \right) \right) - r_{k,c} .$$

Taking the expectation of both sides over $c$, and choosing $x$ to maximize the right hand side, we get
$$\sum_{t=1}^{T} F_t \left( G_{k,c}^t + g_{k,c}^t \right) \geq \max_{x \in P_k \times \{c\}} \left\{ \sum_{t=1}^{T} F_t \left( G_{k,c}^t + x \right) \right\} - \varepsilon_{k,c} \tag{7.3}$$

where we define $F_t(S) = \mathbb{E}_{\vec{c}} [f_t(\mathsf{sample}_{\vec{c}}(S))]$ and $\varepsilon_{k,c} = \mathbb{E}[r_{k,c}]$.

We now define some additional notation. For any set $\vec{S}$ of vectors in $\mathcal{V}^T$, define

$$f(\vec{S}) = \sum_{t=1}^{T} f_t\left(\left\{\vec{a}_t : \vec{a} \in \vec{S}\right\}\right) .$$

Next, for any set $\vec{S}$ of (element vector, color) pairs in $\bigcup_{k=1}^{K}(P_k^T \times \{c\})$, define $\mathsf{sample}_{\vec{c}}(\vec{S}) = \bigcup_{k=1}^{K}\left\{\vec{x} \in P_k^T : (\vec{x}, c_k) \in \vec{S}\right\}$. Define $F(\vec{S}) = \mathbb{E}_{\vec{c}}\left[f(\mathsf{sample}_{\vec{c}}(\vec{S}))\right]$. By linearity of expectation,

$$F(\vec{S}) = \sum_{t=1}^{T} F_t\left(\left\{(\vec{x}_t, c) : (\vec{x}, c) \in \vec{S}\right\}\right) . \tag{7.4}$$

Let $\vec{g}_{k,c} = (\vec{x}, c)$, where $\vec{x}$ is such that $(\vec{x}_t, c) = g_{k,c}^t$ for all $t \in [T]$. Analogously to $G_{k,c}^{t-}$, define $\vec{G}_{k,c}^{-} = \{\vec{g}_{k',c'} : k' \in [K], c' < c\} \cup \{\vec{g}_{k',c} : k' < k\}$. By (7.4), for any $(\vec{x}, c) \in \mathcal{V}^T \times [C]$ we have $F(\vec{G}_{k,c}^{-} + (\vec{x}, c)) = \sum_{t=1}^{T} F_t(G_{k,c}^{t-} + (\vec{x}_t, c))$. Combining this with (7.3), we get

$$F\left(\vec{G}_{k,c}^{-} + \vec{g}_{k,c}\right) \geq \max_{a: a \in P_k}\left\{F\left(\vec{G}_{k,c}^{-} + (a^T, c)\right)\right\} - \varepsilon_{k,c} \tag{7.5}$$

where $a^T$ is the unique element of $\{a\}^T$. Having proved (7.5), we can now use Theorem 9 to complete the proof. Let $\vec{P}_k := \{a^T : a \in P_k\}$ for each $k \in [K]$, and define a new partition matroid over ground set $\{a^T : a \in \mathcal{V}\}$ with feasible solutions $\vec{\mathcal{P}} := \left\{\vec{S} : \forall k \in [K], |\vec{S} \cap \vec{P}_k| \leq 1\right\}$. Let $\vec{G} = \{\vec{g}_{k,c} : k \in [K], c \in [C]\}$. As argued in the proof of Lemma 4, $F_t$ is monotone submodular. Using this fact together with (7.4), it is straightforward to show that $F$ itself is monotone submodular. Thus by Theorem 9,

$$F(\vec{G}) \geq \beta(C, K) \cdot \max_{\vec{S} \in \vec{\mathcal{P}}}\left\{f(\vec{S})\right\} - \sum_{k=1}^{K}\sum_{c=1}^{C} \varepsilon_{k,c} .$$

To complete the proof, it suffices to show that $F(\vec{G}) = \mathbb{E}\left[\sum_{t=1}^{T} f_t(G_t)\right]$, and that $\max_{\vec{S} \in \vec{\mathcal{P}}}\left\{f(\vec{S})\right\} \geq \max_{S \in \mathcal{P}}\left\{\sum_{t=1}^{T} f_t(S)\right\}$. Both facts follow easily from the definitions. □